\documentclass{article}

    \usepackage[preprint]{neurips_2024}

\usepackage[utf8]{inputenc} 
\usepackage[T1]{fontenc}    
\usepackage{url}            
\usepackage{booktabs}       
\usepackage{amsfonts}       
\usepackage{nicefrac}       
\usepackage{microtype}      
\usepackage{xcolor}         

\usepackage[utf8]{inputenc}
\usepackage{amsmath}
\usepackage{amsfonts}
\usepackage{amssymb}
\usepackage{color,xcolor}
\usepackage{comment}
\usepackage{algorithmicx}
\usepackage{algorithm}
\usepackage{algpseudocode}
\usepackage{graphicx}
 \usepackage{booktabs}
 \usepackage{siunitx}
 \usepackage{booktabs}

\usepackage{cite}
\usepackage{mathabx}

\DeclareMathOperator*{\argmin}{arg\,min}
\newcommand{\R}{\mathbb{R}}

\newcommand{\rx}{\mathrm{x}}
\newcommand{\ry}{\mathrm{y}}
\newcommand{\rz}{\mathrm{z}}
\newcommand{\rs}{\mathrm{s}}

\newcommand{\x}{\rx}
\newcommand{\y}{\ry}
\newcommand{\z}{\rz}
\newcommand{\s}{\rs}

\newcommand{\Si}{\mathcal{S}^{\kappa}}

\newcommand{\w}{\mathbf{w}}

\newcommand{\A}{\mathbf{A}}

\newcommand\norm[1]{\left\lVert#1\right\rVert}
\newcommand{\vv}{\mathbf{v}}
\newcommand{\bb}{\mathbf{b}}
\newcommand{\uu}{\mathbf{u}}

\newcommand{\st}{\text{subject to }}
\newcommand{\Rn}{\mathbb{R}^n}

\def\er/{Erd\H{o}s-R\'enyi}

\newtheorem{theorem}{Theorem}[section]
 
\DeclareMathOperator{\prox}{prox}

\newcommand\oprocendsymbol{\hbox{$\blacksquare$}}
\newcommand\oprocend{\relax\ifmmode\else\unskip\hfill\fi\oprocendsymbol}

\title{A GPU-Accelerated Bi-linear ADMM Algorithm for Distributed Sparse Machine Learning}

\author{%
  Alireza Olama
    \\
  Department of Information Technology\\
  \AA{}bo Akademi University\\
  Vaasa, Finland \\
  \texttt{Alireza.Olama@abo.fi} \\
  \And
  Andreas Lundell \\
  Department of Information Technology\\
  \AA{}bo Akademi University\\
  Vaasa, Finland \\
  \texttt{Andreas.Lundell@abo.fi} \\
  \AND
  Jan Kronqvist \\
  Department of Mathematics\\
  KTH Royal Institute of Technology\\
  Stockholm, Sweden \\
  \texttt{jankr@kth.se} \\
  \And
  Elham Ahmadi \\
  School of Technology and Innovations, Computer Science\\
  University of Vaasa\\
  Vaasa, Finland \\
  \texttt{Elham.Ahmadi@uwasa.fi} \\
  \And
  Eduardo Camponogara \\
   Department of Automation and Systems Engineering\\
  Federal University of Santa Catarina\\
  Florianópolis, Brazil \\
  \texttt{Eduardo.Camponogara@ufsc.br} \\
}

\begin{document}

\maketitle

\begin{abstract}
This paper introduces the Bi-linear consensus Alternating Direction Method of Multipliers (Bi-cADMM), aimed at solving large-scale regularized Sparse Machine Learning (SML) problems defined over a network of computational nodes. Mathematically, these are stated as minimization problems with local convex loss functions over a global decision vector, subject to an explicit $\ell_0$ norm constraint to enforce the desired sparsity. The considered SML problem generalizes different sparse regression and classification models, such as sparse linear and logistic regression, sparse softmax regression, and sparse support vector machines. Bi-cADMM leverages a bi-linear consensus reformulation of the original non-convex SML problem and a hierarchical decomposition strategy that divides the problem into smaller sub-problems amenable to parallel computing. In Bi-cADMM, this decomposition strategy is based on a two-phase approach. Initially, it performs a sample decomposition of the data and distributes local datasets across computational nodes. Subsequently, a delayed feature decomposition of the data is conducted on Graphics Processing Units (GPUs) available to each node. This methodology allows Bi-cADMM to undertake computationally intensive data-centric computations on GPUs, while CPUs handle more cost-effective computations. The proposed algorithm is implemented within an open-source Python package called Parallel Sparse Fitting Toolbox (PsFiT), which is publicly available. Finally, computational experiments demonstrate the efficiency and scalability of our algorithm through numerical benchmarks across various SML problems featuring distributed datasets.
\end{abstract}

\section{Introduction}
In recent years, the exponential growth of data has posed significant challenges in both the design and training of Machine Learning (ML) models \citet{verbraeken2020survey,bertsimas2021sparse}.
Traditional approaches struggle to handle the increasing scale and complexity of datasets, leading to issues of computational inefficiency, overfitting, and lack of interpretability \citet{bertsimas2022sparse,olama2023sparse}. Learning \textit{sparse} and \textit{interpretable} ML models can be beneficial to address these challenges \citet{olama2022distributed,bertsimas2016best,bertsimas2021sparseConvex,bertsimas2020sparse,bertsimas2017logistic,bertsimas2022sparse,bai2016splitting}. 

Training sparse models pose a significant challenge, particularly in scenarios involving decentralized data storage systems, where Distributed Data-Parallel (DDP) training plays a vital role \citep{liu2022distributed}, or within the framework of Federated Learning (FL), where preserving data privacy and conducting decentralized computations are paramount concerns \citep{mcmahan2017communication}. Furthermore, the mathematical optimization problems inherent in SML training tend to be non-convex, as they involve satisfying an $\ell_0$ norm constraint, resulting in a solution space that is a union of finitely many subspaces \citep{tillmann2021cardinality,sun2013recent}.
Numerous studies propose distributed sparse training algorithms through convex relaxations of the original non-convex problem. One popular method is distributed Lasso, where an $\ell_1$ norm relaxation is used (see \citet{boyd2011distributed} and the references therein). However, despite the favorable computational properties of Lasso, this method can have some shortcomings, which are well-studied in the statistics community \citet{bertsimas2021sparse,bertsimas2016best}. For example, finding the correct sparsity pattern for general learning problems with the $\ell_1$ norm penalty cannot be guaranteed.

A few papers in the literature address the $\ell_0$ norm-induced sparsity in DDP and FL scenarios. For example, \citet{tong2022federated} presented a federated iterative hard thresholding method for sparse FL in decentralized settings.
In \citet{olama2022distributed}, the authors propose a Distributed Primal Outer Approximation (DiPOA) algorithm designed for SML problems featuring a separable structure. DiPOA is designed based on the outer approximation algorithm \citet{duran1986outer} and the Relaxed Hybrid Alternating Direction Method of Multipliers (RH-ADMM) \citet{olama2019relaxed}. By integrating RH-ADMM into the OA framework, DiPOA distributes the computational load across multiple processors, rendering it suitable for contemporary multi-core architectures.
A recent contribution \citet{olama2023sparse} introduces the Distributed Hybrid Outer Approximation (DiHOA) algorithm, which utilizes a branch-and-bound algorithm tailored for the SML problem structure. DiHOA serves as the core algorithm in the Sparse Convex Optimization Toolkit (SCOT) proposed in \citet{olama2023sparse}.

In both papers provided by \citet{olama2022distributed,olama2023sparse}, the $\ell_0$ norm constraint is addressed by reformulating the original problem into a constraint-coupled Mixed Integer Programming (MIP) equivalent. Although this strategy can solve SML problems to global optimality and satisfy the data privacy constraints, the MIP-based methods face challenges when dealing with large-scale datasets, especially in high-dimensional regimes and distributed settings.
In \citet{olama2023tracking}, the authors propose a distributed algorithm formulated upon a distributed gradient tracking \citet{carnevale2023triggered} and Block Coordinated Descent (BCD) methods. This paper addresses sparsity constraints by iteratively projecting dense solutions onto a set defined by the $\ell_0$ norm. This method is scalable as it avoids adding binary variables to the optimization problem. However, when processing large datasets many iterations are usually needed leading to high communication costs. \citet{fang2020newton}.

In this paper, we propose a scalable algorithm to solve the SML problems distributedly while satisfying data privacy among network nodes. Our contribution lies in developing a distributed algorithm called Bi-Linear consensus ADMM (Bi-cADMM), specifically designed for SML problems featuring $\ell_0$ norm constraints.
The Bi-cADMM algorithm adopts a bi-linear approach to reformulate the problem where the $\ell_0$ norm constraint is exactly reformulated as a bi-linear equality constraint and three convex constraints. Additionally, Bi-cADMM employs a hierarchical decomposition strategy to split the reformulated problem into smaller sub-problems, each of which is private to each computational node. Initially, it conducts a sample decomposition of the original dataset and distributes the local datasets across computational nodes. Following this decomposition, Bi-cADMM performs a feature decomposition,
of each local dataset on the GPUs available to each node.
This strategy enables Bi-cADMM to execute computationally intensive data-centric computations on GPUs, while CPUs handle more cost-effective computations.
The main contributions of this paper are summarized as follows:

\begin{enumerate}
    \item We introduce the Bi-cADMM algorithm that distributedly solves SML problems induced by the $\ell_0$ norm.
    \item We propose a hierarchical data decomposition approach to make the Bi-cADMM algorithm adaptable across hardware platforms, from single CPU-GPU setups to large distributed networks.
    \item We introduce the PsFit Python package to train SML models with the Bi-cADMM optimizer.
    \item We validate the algorithm's effectiveness through various numerical scenarios.
\end{enumerate}


\section{Problem Description}
The SML problem is defined over a network of $N$ computational nodes as a minimization fitting problem incorporating a $\ell_2$ regularization term \footnote{also known as ridge or Tikhonov regularization} \citet{tikhonov1943stability} and an explicit sparsity constraint defined by the $\ell_0$ norm. Mathematically, it can be represented as:
\begin{equation}
    \begin{aligned} \label{p:SML}
        \min_{\x \in \Rn} &\sum_{i = 1}^N \ell_i(\A_i\x - \bb_i) + \frac{1}{2 \gamma}\norm{\x}_2^2\\
        \st~ &\norm{\x}_0 \leq \kappa,
    \end{aligned}
\end{equation}
where $\x \in \Rn$ is a global decision vector, $\A_i \in \R^{m_i \times n}$ is the local feature matrix, $\bb_i \in \R^{m_i}$ is the local output vector, $\gamma >0$ is the regularization weight that controls the importance of the regularization, and $\ell_i: \Rn\xrightarrow{}\R$ is a local convex loss function. 
%
%
The $\ell_2$ regularization helps to reduce the effect of noise in the input data and improves the model's prediction performance. From a numerical perspective, the introduction of the $\ell_2$ regularization promotes the strong convexity of the objective function, thereby improving the condition number of the objective function's Hessian matrix. 
%
%
By choosing different loss function $\ell_i$, different sparse regression and classification models such as Sparse Linear Regression (SLinR), Sparse Logistic Regression (SLogR), Sparse Softmax Regression (SSR), and Sparse Support Vector Machines (SSVMs) can be obtained.

The separable structure of the minimization problem \eqref{p:SML} can be efficiently exploited to design hierarchical DDP  and FL algorithms decomposing the entire problem into small sub-problems solvable in parallel over a network of computation nodes.

\subsection{Bi-linear Consensus Reformulation}
In this section, we present a bi-linear consensus reformulation of problem \eqref{p:SML}, enabling the development of a distributed algorithm to solve \eqref{p:SML}. Initially, we reformulate the sparsity constraint $\|\mathbf{x}\|_0 = \kappa$ as a set of norm inequality constraints and a single bi-linear equality constraint as presented in the following theorem. 
\begin{theorem}[\citet{hempel2014novel}]
    For any vector $\x \in \Rn$, the condition $\norm{\x}_0 \leq \kappa$ holds if, and only if, a vector $\s \in \Rn$ and a scalar $t \in \R$ exist such that,
    \begin{equation}
        \begin{aligned}
            \x^T\s = t,\quad \norm{\x}_1 \leq t,\quad \norm{\s}_1 \leq \kappa,\quad \norm{\s}_\infty \leq  1
        \end{aligned}
    \end{equation}
\end{theorem}
According to the theorem cited, the $\ell_0$ norm constraint can be transformed into a combination of one bi-linear constraint and three linear constraints. Furthermore, we generate a local estimate of the global decision vector $\x$, denoting it as $\x_i$, and distribute it to each node. Subsequently, a consensus constraint is enforced on the problem, ensuring consensus among all local estimates $\x_i$. Hence, problem \eqref{p:SML} can be expressed equivalently as a consensus bi-linear optimization problem defined as,
\begin{equation}
    \begin{aligned} \label{p:bi-convex}
        \min_{\x,\z, t, \s \in \Si} &\sum_{i = 1}^N \ell_i(\A_i\x_i - \bb_i) + \frac{1}{2 N\gamma}\norm{\x_i}_2^2\\
        \text{subject to}~ & \x_i = \z,~\forall i = 1\dots N\\
        & g(\z, \s, t) = 0\\
        &\norm{\z}_1 \leq t.\\
    \end{aligned}
\end{equation}
Here, $\z$ is the consensus variable, $\Si = \{\s \in \Rn: \norm{\s}_{\infty} \leq 1, \norm{\s}_{1} \leq \kappa\}$, and $g(\z, \s, t) = \z^T\s -t $. It can be readily observed that $g(\z, \s, t)$ is linear if either $\z$ or $\s$ is fixed.

\section{Distributed Bi-Linear ADMM Algorithm}
In this section, we develop the Bi-linear consensus ADMM (Bi-cADMM) algorithm to solve problem~\eqref{p:bi-convex} in a distributed fashion. We first construct the augmented Lagrangian for problem~\eqref{p:bi-convex}, incorporating two penalty terms—one for consensus and another for the bi-linear constraint—as follows: 
\begin{equation}
    \begin{aligned} \label{eq:al}
        L(\x,\z, \y, \lambda, \s, t) &
        = \sum_{i = 1}^N \ell_i(\A_i\x_i - \bb_i) + \frac{1}{2 N \gamma } \norm{\x_i}_2^2 
        + \y_i^T (\z - \x_i) +\frac{\rho_c}{2}\norm{\z - \x_i}_2^2\\
        &\quad + \lambda g(\z, \s, t) + \frac{\rho_b}{2} g(\z, \s, t)^2 = L_{\rho_c}(\x,\z, \y)+L_{\rho_b}(\z,\lambda, \s, t).
    \end{aligned}
\end{equation}

Here, \textcolor{black}{$\y_i$ and $\lambda$ are Lagrange multipliers, $\rho_c > 0$ and $\rho_b>0$ are penalties,} and $L_{\rho_c}$ and $L_{\rho_b}$ represent augmented Lagrangian functions for consensus and bi-linear constraints, defined as:
\begin{equation}
        L_{\rho_c}(\x,\z, \y) = \sum_{i=1}^N L^i_{\rho_c}(\x_i,\z, \y_i)\qquad \text{and} \qquad 
        L_{\rho_b}(\z,\lambda, \s, t) = \lambda g(\z, \s, t) + \frac{\rho_b}{2} g(\z, \s, t)^2,
\end{equation}
where the local augmented Lagrangian function $L^i_{\rho_c}(\x_i,\z, \y_i)$ is expressed as:
\begin{equation}
        L^i_{\rho_c}(\x_i,\z, \y_i)=\ell_i(\A_i\x_i - \bb_i) + \frac{1}{2 N \gamma } \norm{\x_i}_2^2 + \y_i^T (\z - \x_i) 
        +\frac{\rho_c}{2}\norm{\z - \x_i}_2^2.
\end{equation}
Therefore, the ADMM steps at iteration $k + 1$ can be written as
\begin{subequations} \label{iterations}
    \begin{align}
        \x_i^{k + 1} &= \argmin_{\x_i}  L^i_{\rho_c}(    \x_i,\z^k, \y_i^k) \label{x-min-step}\\
        (\z^{k + 1}, t^{k + 1}) &= \argmin_{\norm{\z}_1 \leq t} \sum_{i = 1}^N L^i_{\rho_c}(\x_i^{k+1},\z, \y_i^{k}) + L_{\rho_b} (\z, \lambda^k, \s^k, t) \label{global-z-t-updae} \\ 
        \s^{k + 1} &= \textcolor{black}{\argmin_{\s \in \Si} L_{\rho_b}(\z^{k+1},\lambda^k, \s, t^{k+1})} \label{global-s-updae}\\
        \y_i^{k + 1} &= \y_i^{k} + \rho_c (\x_i^{k+1} - \z^{k + 1}) \label{local-dual-update}\\
        \lambda^{k + 1} &= \lambda^k + \rho_b g(\z^{k+1}, \s^{k+1}, t^{k + 1}). \label{global-lambda-updae}
    \end{align}
\end{subequations}
The minimization \eqref{x-min-step} can be written as,
\begin{equation}
        \x_i^{k + 1} =  \argmin_{\x_i}  \ell_i(\A_i\x_i - \bb_i) + \frac{1}{2 N \gamma } \norm{\x_i}_2^2+\frac{\rho_c}{2}\norm{\x_i - \z^k + \uu_i^k }_2^2,
\end{equation}
where $\uu_i^k =\frac{1}{\rho_c}\y_i^k$ and is updated according to
\begin{align}\label{u-update}
    \uu_i^{k+1} &= \uu_i^k +  \x_i^{k+1} - \z^{k+1}.
\end{align}
where \eqref{local-dual-update} is used. The minimization step \eqref{x-min-step} is reduced to the computation of the proximal operator \citet{parikh2014proximal} of the loss function stated as,
\begin{equation}
    \begin{aligned}\label{p:x-subp}
    \x_i^{k + 1} &=  \prox_{\eta f_i}(\z^{k} - \uu_i^{k})
    \end{aligned}
\end{equation}
\textcolor{black}{where $\prox_{\eta f}(\vv) = \argmin_\x f(\x) + \frac{1}{2\eta}\norm{\x - \vv}_2^2$ with $\eta = \frac{1}{\rho_c}$} and $f_i(\x_i) = \ell_i(\A_i\x_i - \bb_i) + \frac{1}{2 N \gamma } \norm{\x_i}_2^2$. 
Therefore, the minimization step \eqref{p:x-subp} involves solving an unconstrained and possibly high-dimensional regularized strongly convex optimization problem. To utilize parallelism, we split problem \eqref{p:x-subp} across features and process each split on a GPU available in each node as discussed in the next section. The minimization steps \eqref{global-z-t-updae} and \eqref{global-s-updae} are convex quadratic optimization problems performed on a coordinator node. These optimization problems do not depend on the raw data and can be constructed by collecting local parameter estimates $\x_i^{k+1}$ and \textcolor{black}{dual estimates $\y_i^{k}$ from other computational nodes.
The augmented Lagrangian of the bi-linear term, $ L_{\rho_b}(\z,\lambda, \s, t)$, can be written as,
\begin{equation}
    L_{\rho_b}(\z,\lambda, \s, t) = \frac{\rho_b}{2} \left(g(\z, \s, t) + v \right)^2 - \frac{\rho_b}{2} v^2
\end{equation}
where $v = \frac{1}{\rho_b} \lambda$. Hence, the minimization step \eqref{global-s-updae} is written as,
\begin{equation}\label{s-update-final}
    \s^{k + 1} = \argmin_{\s \in \Si} \left(g(\z^{k+1}, \s, t^{k+1}) + v^k \right)^2,
\end{equation}
}  
where $v^k$ is updated by the following iteration,
\begin{equation}\label{v-update}
    v^{k+1} = v^k +  g(\z^{k+1}, \s, t^{k+1}) 
\end{equation}
which is obtained by multiplying both sides of \eqref{global-lambda-updae} to $\frac{1}{\rho_b}$.
\paragraph{Termination}
Similar to the standard distributed ADMM described by \citet{boyd2011distributed}, the convergence of iterations in \eqref{iterations} can be monitored using primal and dual residuals denoted by $p_r^k$ and $d_r^k$, respectively. Additionally, we also consider the residual related to the bi-linear equality constraint denoted by $b_r^k$. These residuals are defined as
\begin{equation}
p_r^k = \sum_{i = 1}^N \norm{x_i^k - z^k}_2, \quad d_r^k=\sqrt{N}\rho_c\norm{z^k - z^{k-1}}, \quad b_r^k = \norm{g(\z^{k}, \s^k, t^{k})}_2.
\end{equation}
Thus, the iterations in \eqref{iterations} can be terminated when $p_r^k$, $d_r^k$, and $b_r^k$ fall within a specified small tolerance. 
\subsection{GPU Accelerated Data-Parallel Sub-Solver}
In this section, we develop an augmented Lagrangian-based algorithm to evaluate the proximal operator \eqref{p:x-subp} across the GPUs available within each computing node. Our method involves breaking down the local dataset into partitions based on features and assigning each partition to a separate GPU where a portion of the model fitting problem is solved. From perspective of node $i$, problem \eqref{p:x-subp} can be seen as:
\begin{equation}\label{p:localprox}
        \min_{\x_i}  \ell_i(\A_i\x_i - \bb_i) + \frac{1}{2 N \gamma } \norm{\x_i}_2^2 +\frac{\rho_c}{2}\norm{\x_i - \z^k + \uu_i^k }_2^2.
\end{equation}
We partition the local parameter vector $\x_i$ into $M$ blocks as $\x_i = \left[\x_{i1},\dots,\x_{iM}\right]$, with $\x_{ij} \in \R^{n_j}$, where $\sum_{j=1}^Mn_j = n_i = n$. Hence, the local feature matrix $\A_i$ can be partitioned as, $\A_i = \left[\A_{i1},\dots,\A_{iM}\right]$ with $\A_{ij} \in \R^{m_i \times n_j}$. Therefore, the local subproblem \eqref{p:localprox} can be written as,
\begin{equation}\label{p:local-column-partitioned}
   \min_{\x_i}~ \ell_i(\sum_{j = 1}^M\A_{ij}\x_{ij} - \bb_i) +\sum_{j = 1}^M \frac{1}{2 N \gamma }  \norm{\x_{ij}}_2^2
       +\frac{\rho_c}{2} \norm{\x_{ij} - \z_j^k + \uu_{ij}^k }_2^2,
    \end{equation}
where $\z^k=\left[ \z_1^k,\dots,\z_M^k\right ]$ and $\uu_{i}^k =\left[\uu_{i1}^k,\dots,\uu_{iM}^k  \right]$ are also split in $M$ blocks.
    By defining $r_j(\x_{ij})$ as
\begin{equation}
       \textcolor{black}{r_j(\x_{ij})} = \frac{1}{2 N \gamma }  \norm{\x_{ij}}_2^2 
       +\frac{\rho_c}{2}\norm{\x_{ij} - \z_j^k + \uu_{ij}^k }_2^2,
\end{equation}
problem \eqref{p:local-column-partitioned} becomes
\begin{equation}
    \begin{aligned}
        \min_{\x_i}~&\ell_i(\sum_{j = 1}^M\omega_j - \bb_i) +\sum_{j=1}^M \textcolor{black}{r_j(\x_{ij})} \\
       \st~&\A_{ij}\x_{ij} = \omega_j,
    \end{aligned}
\end{equation} 
where $\omega_j$ is an auxiliary variable. The Augmented Lagrangian function for this problem now becomes:
\begin{equation}
    L_{\rho_l} = \ell_i(\sum_{j = 1}^M\omega_j - \bb_i) +\sum_{j=1}^M \Bigl\{ r_j(\x_{ij}) + \mu_j^T(\A_{ij}\x_{ij} - \omega_j)
        + \frac{\rho_l}{2}\norm{\A_{ij}\x_{ij} - \omega_j}_2^2 \Bigr \}.
    \end{equation}
The ADMM steps in the scaled form are written as,
\begin{equation}
    \begin{aligned}
        \x_{ij}^{k+1} & = \underset{\x_{ij}}{\argmin} ~r_j(\x_{ij}) + \frac{\rho_l}{2}\norm{\A_{ij}\x_{ij} - \omega_j^k + \nu_j^k}_2^2\\
        \omega^{k+1} &= \argmin_\omega ~ \ell_i(\sum_{j = 1}^M\omega_j - \bb_i)  + \sum_{\textcolor{black}{j=1}}^M\frac{\rho_l}{2} \norm{\A_{ij}\x_{ij}^{k+1} - \omega_j + \nu_j^{k}}_2^2\\
        \nu_j^{k+1} & = \nu_j^{k} + A_{ij}\x_{ij}^{k+1} - \omega_j^{k+1},
    \end{aligned}
\end{equation}
where $\nu_j^k = \frac{1}{\rho_l}\mu_j^k$. Each GPU within a node is responsible for solving the first and last steps independently in parallel for each $j=1,\dots,M$. The $\omega$-update requires solving a problem in $Mn_j$ variables, however,
it is shown in \citet{boyd2011distributed} that
this step can be carried out by solving a problem only in $n_j$ variables by optimizing over $\widebar{\omega}_j = \frac{1}{M}\omega_j$, \emph{i.e.,} the average of all local $\omega_j$ variables. The $\widebar{\omega}$-minimization step is written as,
\begin{equation}\label{p:GPU;omega}
        \widebar{\omega}^{k+1} = \argmin_{\widebar{\omega}}\ell_i(M\widebar{\omega} - \bb_i) 
        + \frac{M\rho_l}{2}\norm{\widebar{\omega}- \widebar{\A_{ij}\x_{ij}}^{k+1} - \nu^k}_2^2,
\end{equation}
where $\widebar{\A_{ij}\x_{ij}}^{k+1} = \frac{1}{M}\sum_{j=1}^M\A_{ij}\x_{ij}^{k+1}$. Similarly, the $\nu_j$-update becomes
\begin{equation}
    \begin{aligned}\label{p:GPU;nu}
        \nu^{k+1} & = \nu^{k} + \widebar{\A_{ij}\x_{ij}}^{k+1} - \widebar{\omega}^{k+1}.
    \end{aligned}
\end{equation}
Finally, the $\x_{ij}$-minimization step is written as
\begin{equation}\label{p:GPU;x}
        \x_{ij}^{k+1}  = \underset{\x_{ij}}{\argmin} ~ r_j(\x_{ij})
         +\frac{\rho_l}{2}\norm{\A_{ij}\x_{ij} - \A_{ij}\x_{ij}^k - \widebar{\omega}^k + \widebar{\A_{ij}\x_{ij}}^{k} + \nu^k}_2^2.
\end{equation}
The $\x_{ij}$-minimization step involves solving $M$ parallel regularized least-squares problems, each with $n_j$ variables. Following this, we aggregate and sum the partial predictors $\A_{ij}\x_{ij}^{k+1}$ between the first and second steps to construct $\widebar{\A_{ij}\x_{ij}}^{k+1}$. The second stage involves a single minimization problem in $m_i$ variables, focusing on quadratically regularized loss minimization. Subsequently, the third stage comprises a straightforward update in $m_i$ variables. Since the loss function $\ell_i$ is separable, the $\Bar{\omega}$-update splits entirely into $m_i$ scalar optimization problems.
\begin{algorithm}
\begin{algorithmic}
    \State \textbf{Initialization:} Initialize $\uu_i^0$, $\x_i^0$ for $i = 1,\dots, N$
    \For{$k=0,\dots,K$}
    \State \textbf{Collect:} Gather $\x_i^k$ and $\uu_i^k$ from all nodes $i = 1,\dots, N$
    \State \textbf{Global Updates:}
    \State $\quad$ Solve \eqref{global-z-t-updae} to compute $\z^{k+1}$ and $t^{k+1}$
    \State $\quad$ Solve \eqref{s-update-final} to update $\s^{k+1}$
    \State $\quad$ Update $v^k$ using \eqref{v-update}
    \State \textbf{Bcast:} broadcast $\z^{k+1}$ to all nodes $i = 1,\dots, N$
    \For{nodes $i = 1,\dots, N$} in \textbf{parallel}
        \State \textbf{Wait:} Receive $\z^{k+1}$ from the global node
        \State Update $\uu_i^{k+1}$ from \eqref{u-update}
        \State \textbf{Partition:} Split $\z^{k+1}$ and $\uu_i^{k+1}$ into $M$ partitions
        \State \textbf{Send:} Transfer $\z^{k+1}_j$ and $\uu_{ij}^{k+1}$ to $j$-th GPU of the $i$-th node
        \For{GPUs $j = 1,\dots, M$} in \textbf{parallel}
        \State Update $\x_i^{k+1}$ using Algorithm 2.
        \EndFor
    \EndFor
    \EndFor
\end{algorithmic}
\caption{Bi-cADMM Algorithm \label{global-algorithm}}
\end{algorithm}
\subsection{Distributed Implementation and Bi-cADMM Pseudo-Code}
This section describes the algorithm's pseudo-code corresponding to the main steps discussed earlier. We categorize the algorithm into three primary updates: \textit{global-level}, \textit{node-level}, and \textit{device-level}.
The global-level update encompasses steps executed on the CPU of the global node with network-level communications. These computations are independent of problem-specific data such as $\A_i$ and $\bb_i$. The node-level computations are primarily data-driven and are carried out on the GPUs available to each node with possible inter-GPU communications. Consequently, data partitions like $\A_{ij}$ reside on the $j$-th GPU of the $i$-th node during algorithm execution. Lastly, the device-level computations involve solving the sub-problem \eqref{p:GPU;x} on each GPU (or CPU in case GPUs are not available) within a node, primarily focusing on thread-level parallel computations. The pseudo-code of Bi-cADMM and the node-level algorithms are presented in Algorithms \ref{global-algorithm} and \ref{alg-local}, respectively.

 Within each Bi-cADMM iteration, there are several key stages. First, all nodes gather the current values of their variables. Then, a global update is performed, involving solving optimization problems to compute global variables such as $\z$ and $\lambda$ which are broadcast to all nodes. Subsequently, each node receives the updated global variables and proceeds with its computations. This involves updating its local variables, partitioning the global variables for efficient GPU processing, and distributing the computation across multiple GPUs within each node. Finally, each node updates its local variables using its allocated GPUs in parallel. 
\begin{algorithm}
\begin{algorithmic}
    \State \textbf{Initialization:} $\nu^0$
    \While{\textbf{termination}}
    \For{$j = 1,\dots, M$} in \textbf{parallel}
    \State \textbf{Collect:} Gather $\z_j^k$ and $\uu_{ij}^k$ from $i$-th node
    \State Compute $\x_{ij}$ from \eqref{p:GPU;x}
    \State Compute $\A_{ij}\x_{ij}$
    \State $\w = \A_{ij}\x_{ij}$
    \EndFor
    \State \textbf{Allreduce} $\w$ and compute $\widebar{\A_{ij}\x_{ij}}$
    \State Compute $\Bar{\omega}$ by solving \eqref{p:GPU;omega}
    \State Update $\nu$ by using \eqref{p:GPU;nu}
    \EndWhile
\end{algorithmic}
\caption{Node level updates\label{alg-local}}
\end{algorithm}

In the node-level algorithm, each GPU unit gathers $\z_j^k$ and $\uu_{ij}^k$ from the corresponding node. Then, it computes $\x_{ij}$ using an equation \eqref{p:GPU;x}. Following this, it computes $\A_{ij}\x_{ij}$, denoted as $\w$, representing a matrix-vector product, independently across all GPUs in parallel. Subsequently, the computed $\w$ values are aggregated across all GPUs using an \texttt{AllReduce} operation to compute $\widebar{\A_{ij}\x_{ij}}$. 


\section{Numerical Experiments} 
\begin{figure}
    \centering
    \includegraphics[width=1\textwidth]{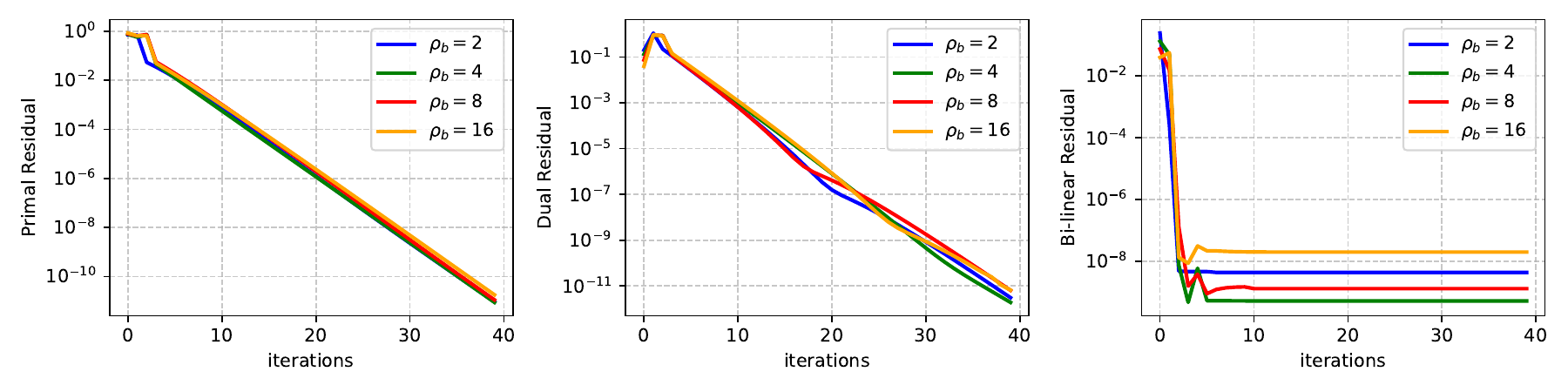}
    \caption{\footnotesize Primal, dual, and bi-linear residuals for different bi-linear penalty parameters, $\rho_b = 2, 4, 8, 16$.}
    \label{fig:conv}
\end{figure}

In this section, we present the performance results of applying Algorithm \ref{global-algorithm} across various numerical scenarios. Our investigation revolves around assessing the algorithm's performance in training SLS models using synthetic datasets generated according to ground truth models that are known to be sparse. The SLS problem is defined as the following minimization problem,
\begin{equation}
    \begin{aligned} \label{p:sls}
        \min_{\x \in \Rn} &\sum_{i = 1}^N \norm{\A_i\x - \bb_i}_2^2 + \frac{1}{2 \gamma}\norm{\x}_2^2\\
        \st~ &\norm{\x}_0 \leq \kappa.
    \end{aligned}
\end{equation}
We evaluate the performance of Algorithm \ref{global-algorithm} across four scenarios. In the first scenario, we evaluate the empirical convergence of the algorithm for different $\rho_b$. In the second scenario, we compare the solution time of the Bi-cADMM algorithm with two other methods: an exact MIP reformulation, as described by \citet{olama2023sparse}, \citet{bertsimas2017logistic}, and \citet{bertsimas2016best}, solved using Gurobi \citet{gurobi} 11.0 with an academic license, and the Lasso method \citet{tibshirani1996regression}. In the third and fourth scenarios, we assess the algorithm's scalability concerning the number of features and data points. 
\paragraph{Hardware and Software}
All the experiments were performed on a machine running Ubuntu 22.04 equipped with an Intel(R) Core i7-13700 processor clocked at 2.1 GHz, featuring 16 physical cores and 32 GB memory. The execution of Algorithm \ref{alg-local} was performed on an NVIDIA GeForce RTX 4070 GPU with 12 GB memory.
Algorithms \ref{global-algorithm} and \ref{alg-local} are implemented within the PsFit package which is entirely written in Python 3.11.
%
For performing the linear algebra operations required by the algorithms, \texttt{PyTorch} 2.3.0 with Cuda support was employed. Distributed computing operations, including \texttt{AllReduce} and \texttt{BCast}, were implemented through the utilization of \texttt{mpi4py} which is Python's Message Passing Interface (MPI) library.
We consider dense local feature matrices, denoted as $\A_i$, for all nodes, where the elements are drawn from a standard normal distribution. Subsequently, we normalize the columns of $\A_i$ to have unit $\ell_2$ norms. A ground truth vector, $\x_{\mathrm{true}} \in \R^n$, is generated with a given sparsity level parameter, $0 < s_l < 1$, dictating the degree of sparsity in the true solution to be retrieved by Algorithm \ref{global-algorithm}. Here, the count of nonzero elements used in the algorithm is calculated as $\kappa = \text{round}(n(1 - s_l))$. Local labels, denoted as $\bb_i$, are computed as $\bb_i = \A_i\x_{\mathrm{true}} + \mathbf{e}$, where $\mathbf{e}$ is sampled from a normal distribution $\mathcal{N}(0, \Sigma)$. Subsequently, the locally processed datasets are transmitted to each respective node for further data processing.
\paragraph{Empirical Convergence}
This section evaluates the empirical convergence of Algorithm \eqref{global-algorithm}. The primary metrics for evaluating convergence are the primal and dual residuals, along with bi-linear residuals, all on a logarithmic scale. Figure \ref{fig:conv} illustrates these residuals for various bi-linear penalty parameters $\rho_b$. Our experiments indicate that $\rho_b$ should be chosen relative to $\rho_c$ such that $\rho_b \leq \alpha \rho_c$, where $\alpha$ is in the range (0,1]. This selection strategy ensures that the algorithm reaches consensus before satisfying the bi-linear constraint, which empirically results in better convergence in practice. For this study, we set $n = 4000$, $m = 10000$, $s_l = 0.8$, and $\alpha = 0.5$. As shown in Figure \ref{fig:conv}, while $\rho_b$ has a minimal impact on the primal and dual residuals, it significantly influences the convergence of the bi-linear residuals, as expected.

\begin{table}[ht]
    \centering
    \caption{\footnotesize Comparison of solution time between Bi-cADMM, Gurobi, and Lasso for different $s_l$, $m$, $n$. \label{table: comparison}}
    \small
    \begin{tabular}{cccccccc}
        \toprule
        & & \multicolumn{2}{c}{Bi-cADMM [s]} &\multicolumn{2}{c}{Gurobi [s]} & \multicolumn{2}{c}{Lasso [s]} \\
        \cmidrule{3-4} \cmidrule{5-6} \cmidrule{7-8}
        & & $n = 2k$ & $n = 4k$ &$n = 2k$ & $n = 4k$ & $n = 2k$ & $n = 4k$ \\
        \midrule
        $s_l = 0.6$ & $m = 1 \times 10^5$ & \multicolumn{1}{r}{1.1} & \multicolumn{1}{r}{1.8} & \multicolumn{1}{r}{374.5} & \multicolumn{1}{r}{cut off} & \multicolumn{1}{r}{2.7\textsuperscript{*}} & \multicolumn{1}{r}{8.9\textsuperscript{*}} \\
        & $m = 2 \times10^5$ & \multicolumn{1}{r}{1.7} & \multicolumn{1}{r}{3.0} & \multicolumn{1}{r}{313.2} & \multicolumn{1}{r}{cut off} & \multicolumn{1}{r}{5.5\textsuperscript{*}} & \multicolumn{1}{r}{19.2\textsuperscript{*}} \\
        & $m = 3 \times10^5$ & \multicolumn{1}{r}{2.2} & \multicolumn{1}{r}{4.0} & \multicolumn{1}{r}{300.1} & \multicolumn{1}{r}{cut off} & \multicolumn{1}{r}{7.3\textsuperscript{*}} & \multicolumn{1}{r}{77.6\textsuperscript{*}} \\
        \midrule
        $s_l = 0.9$ & $m = 1 \times10^5$ & \multicolumn{1}{r}{1.1} & \multicolumn{1}{r}{1.9} & \multicolumn{1}{r}{250.5} & \multicolumn{1}{r}{cut off} & \multicolumn{1}{r}{2.7\textsuperscript{*}} & \multicolumn{1}{r}{9.3\textsuperscript{*}} \\
        & $m = 2 \times10^5$ & \multicolumn{1}{r}{1.7} & \multicolumn{1}{r}{2.9} & \multicolumn{1}{r}{241.8} & \multicolumn{1}{r}{cut off} & \multicolumn{1}{r}{5.0\textsuperscript{*}} & \multicolumn{1}{r}{17.4\textsuperscript{*}} \\
        & $m = 3 \times10^5$ & \multicolumn{1}{r}{2.2} & \multicolumn{1}{r}{4.1} & \multicolumn{1}{r}{230.2} & \multicolumn{1}{r}{cut off} & \multicolumn{1}{r}{7.4\textsuperscript{*}} & \multicolumn{1}{r}{70.1\textsuperscript{*}} \\
        \bottomrule
    \end{tabular}
\end{table}

\begin{figure}
    \centering
    \includegraphics[width=1\textwidth]{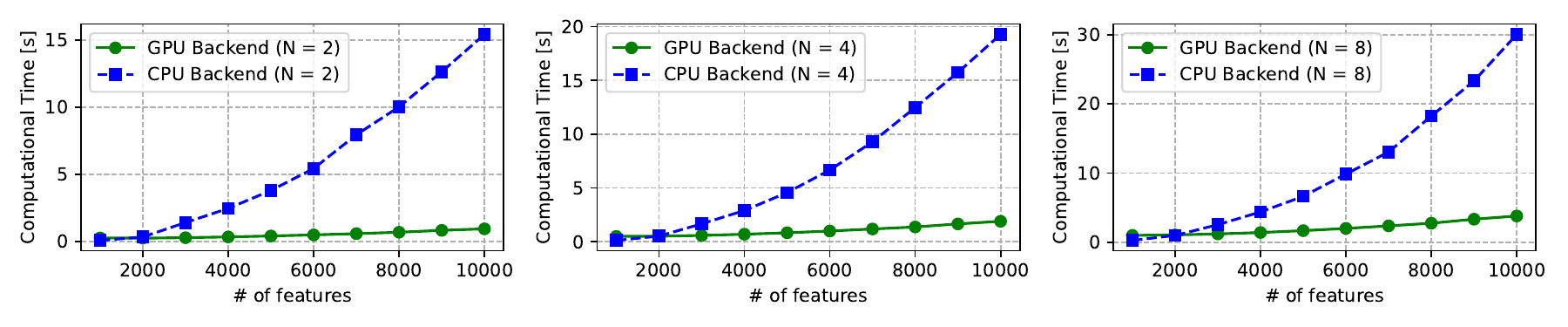}
    \caption{\footnotesize Comparison of computational times for feature scaling scenario across varying numbers of computational nodes ($N = 2, 4, 8$) using both GPU and CPU backends.}
    \label{fig:feature_scaling}
\end{figure}

\paragraph{Computational Time Comparison}
Here we provide the computational results compared to Gurobi and Lasso. We execute the Bi-cADMM algorithm within a network of $N = 4$ nodes each of which contains $\frac{m}{N}$ data points and $n$ features. We run Gurobi with default settings and with a time limit of $1800$ seconds. The Lasso method is implemented using \texttt{glmnet} package with a Python interface available at \url{https://github.com/civisanalytics/python-glmnet.git}.
Table \ref{table: comparison}  presents a comparison of solution times between Bi-cADMM, Gurobi, and Lasso algorithms for varying values of \( s_l \) (sparsity level), \( m \) (number of samples), and \( n \) (number of features). It is evident in the table that Bi-cADMM shows superior performance over the other two methods across all tested scenarios. As the problem size increases, both in terms of sample size and feature count, Bi-cADMM maintains its efficiency, whereas Gurobi often fails to produce results within the cut-off time for larger datasets. Computationally, Lasso performs better than Gurobi,  but still lags behind Bi-cADMM, especially as the problem size grows. The asterisk in the table shows the scenarios in which Lasso could not recover the true sparsity.
\paragraph{Scalability across features}
In this section, we examine the scalability of the Bi-cADMM algorithm as the number of features $n$ increases while the number of data points of each node is fixed. In particular, we consider problem \eqref{p:sls} with $N = 2, 4, 8$ nodes each of which consists of a feature matrix $\A_i \in \R^{800 \times n}$ where $n$ varies from  $1000$ to $10000$. Therefore, the total number of data points for each $N$ is $1600, 3200, 6400$, respectively. In this scenario, the sparsity level is fixed to $s_l = 0.8$ which assumes $80\%$ of sparsity on the model's parameters.
Figure \ref{fig:feature_scaling} presents the performance comparison of Algorithm \ref{global-algorithm} using CPU and GPU backends as the number of features increases. The line graphs depict computational times for feature scaling on CPUs (green lines) and GPUs (blue lines) across different numbers of nodes. It is evident that GPUs consistently outperform CPUs, maintaining lower computational times even as the number of features rises to $10000$. This performance gap suggests that GPUs are more suitable in managing high-dimensional feature matrices.
\paragraph{Scalability across data points}
\begin{figure}
    \centering
    \includegraphics[width=1\textwidth]{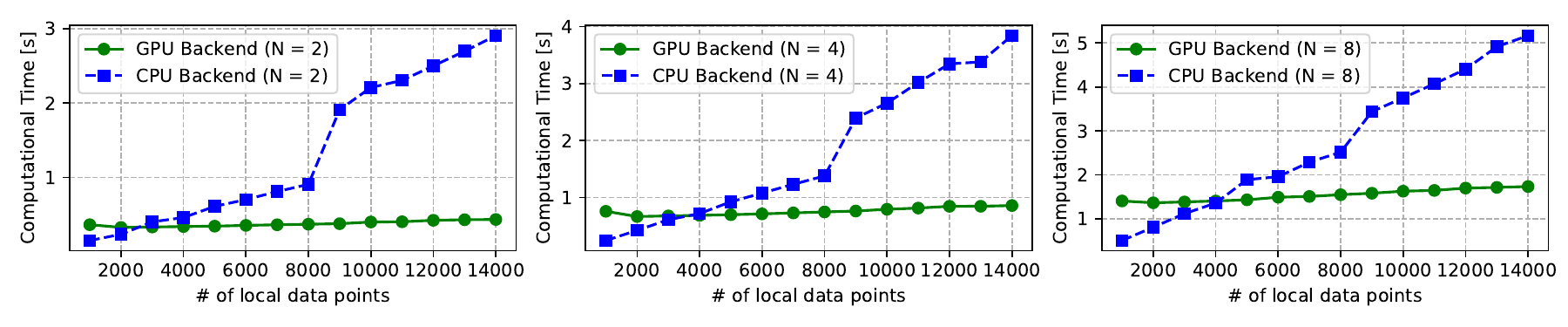}
    \caption{\footnotesize Comparison of computational times for sample scaling scenario across varying numbers of computational nodes ($N = 2, 4, 8$) using both GPU and CPU backends.}
    \label{fig:sample_scaling}
\end{figure}
This section discusses the scalability of Algorithm \ref{global-algorithm} as the number of local data points increases. In this scenario, we fix the number of features and the sparsity level parameter to $n = 4000$ and $s_r = 0.8$ respectively, and increase the number of data points available to each node from $25\times10^3$ to $300\times10^3$. Therefore, the total maximum number of data points for different values of $N$ are $600\times10^3$, $1.2\times10^6$, and $2.4\times10^6$, respectively.
The computational results of this scenario are presented in Figure \ref{fig:sample_scaling}. The computational time for both CPU and GPU backends increases with the number of data points. However, the GPU backend demonstrates a more gradual increase in computational time compared to the CPU backend, which exhibits a steeper climb. This indicates that GPUs handle scaling with larger datasets more efficiently than CPUs, which is consistent with the trend observed in Figure \ref{fig:feature_scaling} regarding the number of features.

\begin{figure}
    \centering
    \includegraphics[width=1\textwidth]{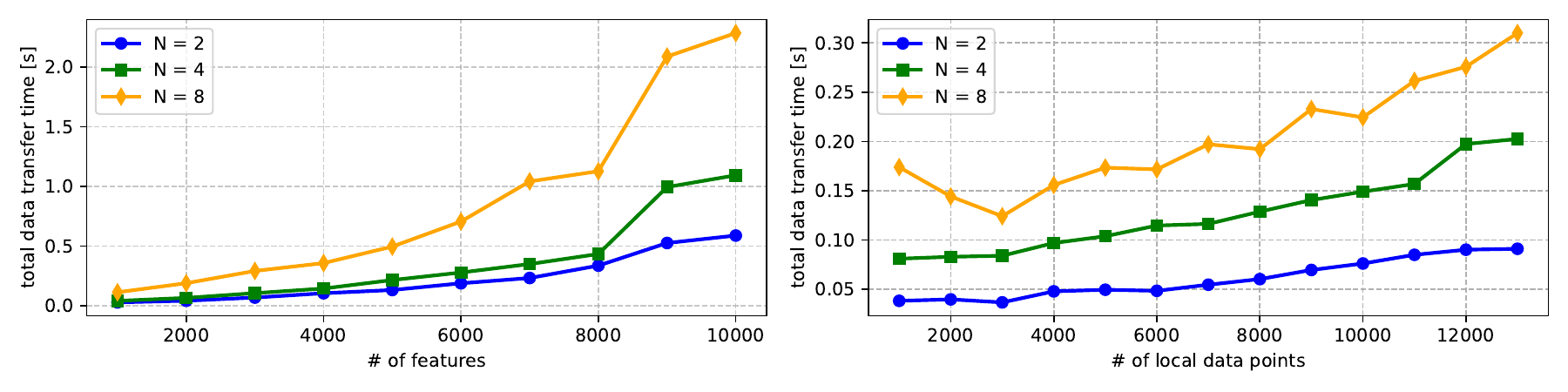}
    \caption{\footnotesize Comparison of total data transfer times for feature and sample scaling scenario across varying numbers of computational nodes ($N = 2, 4, 8$).}
    \label{fig:transfer_time}
\end{figure}
\paragraph{CPU-GPU Memory Transfer}
Here, we evaluate the time spent by the algorithm to transfer data from CPU to GPU and vice versa. In this scenario, we use the same settings as for the previous scenario, however, the total data transfer time during the execution of the algorithm is presented which is shown in Figure \ref{fig:transfer_time}.
In this figure, the solid lines represent different numbers of nodes and the y-axis represents the data transfer time. The computational results indicate that as the number of features or data points increases, the time required for data transfer also increases. However, in the data points scalability scenario less data transfer time is required since the number of features is fixed and at each iteration, a fixed number of parameters is transferred between CPUs and GPUs. This is consistent with the expectation that more data requires more time to move between nodes. Moreover, these results suggest that optimizing data transfer is crucial, especially when dealing with large feature sets or datasets, to ensure efficient use of computational resources. 

\section{Acknowledgment}
This research was funded by Högskolestiftelsen i Österbotten. The authors gratefully acknowledge additional support from FAPESC (grant 2021TR2265), CNPq (grants 308624/2021-1 and 402099/2023-0), Digital Futures, and the C3.ai Digital Transformation Institute.

\section{Conclusions}
In this paper, we introduced the Bi-cADMM algorithm for solving regularized SML problems over a network of computational nodes. Bi-cADMM leverages a bi-linear consensus reformulation of the original SML problem and hierarchical decomposition to enable parallel computing on both CPUs and GPUs. Implemented within an open-source Python package, our algorithm demonstrates efficiency and scalability through numerical benchmarks across various SML problems with distributed datasets. However, the main limitations lie in the necessity of a global node and the lack of convergence proof.  Future work will address these issues by focusing on convergence analysis and optimizing Bi-cADMM for efficient use in multi-GPU and multi-CPU environments.
\bibliographystyle{abbrvnat}
\bibliography{refs}
\end{document}